\documentclass[conference]{IEEEtran}
\IEEEoverridecommandlockouts
\usepackage{algorithm}
\usepackage{algpseudocode}
\usepackage{cite}
\usepackage{graphics}
\usepackage{array}
\usepackage{xcolor}
\usepackage{float}
\usepackage{amsfonts}
\usepackage{import}
\usepackage{tikz}
\usepackage{tabularx}
\usepackage{placeins}
\definecolor{darkpink}{RGB}{199,21,133}
\newcommand{\added}[1]{#1}
\usepackage[colorlinks = true,
            linkcolor = blue,
            urlcolor  = blue,
            citecolor = blue,
            anchorcolor = blue]{hyperref}

\newcommand{\rev}[1]{#1}
\usetikzlibrary{shapes,arrows}
\usepackage{amsthm,amsmath,amssymb,amsfonts}
\usepackage{graphicx}

\usepackage{xpatch}
\makeatletter
\xpatchcmd{\@thm}{\thm@headpunct{.}}{\thm@headpunct{}}{}{}
\makeatother
\usepackage{multirow}
\newcolumntype{L}{>{\centering\arraybackslash}m{0.7cm}}
\graphicspath{{figures/}}
\usepackage{graphicx, pifont} 

\usetikzlibrary{arrows.meta}
\usetikzlibrary{positioning,fit,backgrounds}
  
\tikzset{%
  >={Latex[width=2mm,length=2mm]},
            base/.style = {rectangle, rounded corners, draw=black,
                           minimum width=1.5cm, minimum height=0.7cm,
                           text centered, font=\sffamily\small},
  activityStarts/.style = {base, fill=black!30},
       startstop/.style = {base, fill=cyan!10},
    activityRuns/.style = {ellipse, minimum width=1cm, minimum height=0.7cm, text centered, font=\sffamily\small, fill=red!25},
         process/.style = {base, minimum width=3cm, fill=yellow!30,
                           font=\sffamily\small},
            a/.style = {rectangle, fill = cyan!10
                           minimum width=2cm, minimum height=0.5cm,
                           text centered, font=\sffamily\small},
}

\begin{document}

\title{Rethinking Neural Width for Alternating Current Optimal Power Flow Proxies}
\author{\IEEEauthorblockN{Dhruvi Khandelwal${^\dagger} ^1$, Anurag Basistha${^\dagger} ^2$, Ayushi Jolotia$^3$ , and  Parikshit Pareek${^\star} ^3$ \\ $^1$ Department of Electrical Engineering, National Institute of Technology Kurukshetra, India. \\ $^2$ Indraprastha Institute of Information Technology Delhi, India. \\ $^3$Department of Electrical Engineering, Indian Institute of Technology Roorkee, India.}
\thanks{$^\dagger$Equal Contributions. $^\star$ Corresponding Author: \emph{pareek@ee.iitr.ac.in}}
\thanks{\textcolor{black}{This work was supported by the ANRF PM Early Career Research Grant (ANRF/ECRG/2024/001962/ENS) and the IIT Roorkee Faculty Initiation Grant (IITR/SRIC/1431/FIG-101078) along with IIT Roorkee's SPARK Internship Program which supported Dhruvi's internship in the summer of 2025.}}
}

\maketitle
\begin{abstract}
Deep learning proxies for Alternating Current Optimal Power Flow (ACOPF) lack systematic methods for determining architectural size. This paper conducts a constructive thought experiment to answer a fundamental inquiry: \textit{how wide must a neural network be to almost accurately approximate the ACOPF manifold?} We introduce a Loss-Guided Neural Densification (LG-ND) algorithm that incrementally discovers necessary capacity by expanding only when the current deep neural network topology fails to improve further. Empirical results across various IEEE systems show that LG-ND achieves performance parity with literature baselines using up to ten times fewer neurons per layer. Such architectural minimalism is critical for the formal verification required in safety-critical grid operations.
\end{abstract}


\section{Introduction}
The imperative of maintaining power grid reliability requires solving the highly constrained Alternating Current Optimal Power Flow (ACOPF) problem. While ACOPF is essential for economic dispatch, its critical role lies in ensuring system stability by adhering to rigorous physical limits. Inaccurate or slow solutions do not merely incur financial costs; they risk physical limit violations that can jeopardize grid security. \cite{lovett2024_opfdata}


To bridge the gap between AC feasibility and real-time execution, Machine Learning (ML) proxies have emerged as a solution. A proxy or surrogate is a parameterized network like deep neural network which can provide approximate solutions of ACOPF, given the load inputs, in a single forward pass. By learning the mapping of optimization solvers, these proxies allow for rapid contingency screening \cite{guha2019ml_acopf, rahman2020ml_scopf, fioretto2020predicting_acopf, pan2020deepopf}. However, current approaches largely focus on minimizing data loss rather than ensuring the structural reliability of the model itself. A prevailing trend in learning-based ACOPF is the scaling of model capacity (width and depth) to improve accuracy. While effective in data-rich environments, this ``over-parameterization'' presents a liability for power systems \cite{xu2019dynamic_nn_survey}. \added{Table~\ref{tab:nn_architectures_main} highlights that, even for the same IEEE-118 system, existing proxy models adopt widely varying layer widths}.
Large, opaque Deep Neural Networks (DNNs) are resource-intensive to deploy on edge devices and, critically, are difficult to verify. In safety-critical infrastructure, a smaller, interpretable model is inherently more reliable than a massive ``black box" that consumes excessive computational resources.

\textbf{Positioning:} This work frames a fundamental inquiry regarding model capacity: \textit{how wide must a neural network be to reliably approximate the ACOPF manifold?} Rather than defaulting to over-parameterization, we employ Loss-Guided Neural Densification (LG-ND) as a constructive thought experiment to generate \textit{lean} proxies. This architectural minimalism is critical for formal verification, a primary barrier to grid deployment. State-of-the-art verifiers like $\alpha, \beta$-CROWN \cite{wang2021betacrown} utilize branch-and-bound methods where complexity scales aggressively with width, rendering traditional wide networks intractable to certify. By constraining width to strict physical necessity, we ensure the proxy remains amenable to the rigorous safety checks required for secure operations.

\begin{table}[t]
\centering
\caption{Summary of Neural Network Architectures: Width}
\label{tab:nn_architectures_main}
\begin{tabular}{lc}
\hline
\textbf{Model / Paper} & \textbf{Width (Neurons per Layer)} \\
\hline
\cite{meta_opf} IEEE-118 & 256 / 128 / 64 \\
\cite{lei2022data_driven_lagrange_opf} IEEE-118 & 200 / 200 / 200 \\
\cite{pan2023deepopf} IEEE-118 & 256 / 128 \\
\cite{liang2023deepopf_u} IEEE-118 & 1024 / 512 / 256 \\
\cite{chen2025_physics_informed_gradient_acopf} IEEE-118 & 236 / 50 / 236 \\
\cite{huang2022deepopf_v} IEEE-118 & 512 / 256 / 128 \\
\cite{nguyen2025_fsnet} IEEE-118 & 1024 / 1024 / 1024 / 1024 \\
Proposed IEEE-118 & 50 / 50 \\
\hline
\end{tabular}
\vspace{-1em}
\end{table}

\section{Loss-Guided Densification}
In this paper, our target is a standard ACOPF proxy where input is the load vector (real and reactive stacked together) and output is all ACOPF decision variables \cite{park2023self}. 
Let the power network graph be denoted by $\mathcal{G}(\mathcal{V}, \mathcal{E})$ with $N = |\mathcal{V}|$ buses and \added{$\mathcal{E}$ transmission lines}. We define the input feature vector $\mathbf{x} \in \mathbb{R}^{2N}$ as the concatenation of nodal active ($\mathbf{p}_d$) and reactive ($\mathbf{q}_d$) power demands: $\mathbf{x} := [p_{d_1}, \dots, p_{d_N}, q_{d_1}, \dots, q_{d_N}]^\top$. The target output vector $\mathbf{y} \in \mathbb{R}^{M}$\added{where $M$ denotes the total number of decision variables.} consisting of the optimal decision variables required to fully characterize the grid state. This includes the voltage magnitudes \added{$\mathbf{v}$}, where $v_i$ denotes the magnitude at bus $i$, voltage angles $\boldsymbol{\theta}$, and generator active and reactive power generator setpoints ($p_g, q_g$): $\mathbf{y} := [\mathbf{v}^\top, \boldsymbol{\theta}^\top, \mathbf{p}_g^\top, \mathbf{q}_g^\top]^\top$.

The goal of the proxy is to approximate the optimal mapping $\mathcal{M}: \mathbf{x} \mapsto \mathbf{y}^\star$ defined by the underlying ACOPF solver. We seek a parameterized neural network function $f_\phi(\mathbf{x})$ such that $f_\phi(\mathbf{x}) \approx \mathbf{y}^\star$, where $\mathbf{y}^\star$ is the minimizer of the standard non-convex formulation:
\begin{align}
    \mathbf{y}^\star = & \arg \min_{\mathbf{y}} \mathcal{C}(\mathbf{y}) \\
    \text{s.t.} \quad &\mathbf{g}(\mathbf{x}, \mathbf{y}) = 0 ; \\ 
    & \mathbf{h}(\mathbf{x}, \mathbf{y}) \le 0 .
\end{align}

Here, $\mathcal{C}(\mathbf{y})$ represents the generation cost, $\mathbf{g}(\cdot)$ denotes the non-linear AC power flow balance equations (Kirchhoff’s laws), and $\mathbf{h}(\cdot)$ covers the operational inequality constraints, including branch thermal limits and voltage magnitude bounds.
\added {Specifically, $\mathbf{g}(\mathbf{x}, \mathbf{y}) = 0$ encapsulates the AC power flow balance equations at each bus $i \in \mathcal{V}$:}
\begin{align}
    p_{g,i} - p_{d,i} &= \sum_{j \in \mathcal{V}} v_i v_j|Y_{ij}|\cos(\theta_i - \theta_j - \delta_{ij}), \\
    q_{g,i} - q_{d,i} &= \sum_{j \in \mathcal{V}} v_i v_j|Y_{ij}|\sin(\theta_i - \theta_j - \delta_{ij}),
\end{align}
\added{
where $Y_{ij}$ denotes the $(i,j)$-th element of the network admittance matrix, with magnitude $|Y_{ij}|$ and phase $\delta_{ij}$.
The inequality constraints enforce standard operational limits:}
\begin{align}
    p_{g,i}^{\min} & \le p_{g,i} \le 
    p_{g,i}^{\max}, 
    \\
    q_{g,i}^{\min} & \le q_{g,i} \le q_{g,i}^{\max}, \\
    v_{i}^{\min} &\le v_i \le v_{i}^{\max}, 
    \\ |S_{ij}| & \le S_{ij}^{\max}, \ \forall (i,j)\in\mathcal{E},
\end{align}
\added{
where $S_{ij}$ represents the apparent power flow on branch $(i,j)$.
}

To empirically determine the minimal width required to solve this constrained optimization, we invert the standard design process. Instead of starting with a large model and then pruning it, we employ a constructive algorithm that incrementally discovers the necessary capacity by expanding the network only when the current topology fails to resolve the physical constraints.

The proposed LG-ND strategy (Algorithm \ref{alg:fg-adp}) treats the architectural width not as a hyperparameter, but as a dynamic variable dependent on the complexity of the grid physics. We initialize the proxy $f_\phi$ with a minimal capacity, intentionally under-parameterizing the mapping $\mathcal{M}: \mathbf{x} \mapsto \mathbf{y}^*$. In each iteration, the model is trained until convergence. The validation loss $\mathcal{L}$ serves as a quantitative metric for the model's ability to capture the non-convex ACOPF manifold and satisfy physical constraints $\mathbf{g}(\cdot)$ and $\mathbf{h}(\cdot)$. \added{If the validation loss improves compared to the previous snapshot ($\mathcal{L} < \mathcal{L}^\star$), it indicates that the architecture has not yet reached its saturation point. Consequently, we update the best model state and densify the network by expanding hidden layers by $\Delta$ neurons. In our implementation, we used a fixed step size of $\Delta = 10$ neurons per expansion}. This iterative growth continues only until performance saturates or a strict capacity cap is reached, ensuring the final architecture represents the minimal sufficient width required to model the system. To isolate the structural effects of the ADP, \added{ we employ a standard supervised training regime using Mean Squared Error (MSE) loss, consistent with established ACOPF proxy methodologies \cite{ML4OPF2023}. Specifically, we train $f_\phi$ by minimizing
\begin{equation}
\mathcal{L}_{\mathrm{MSE}}
=\frac{1}{|\mathcal{D}|}
\sum_{(\mathbf{x},\mathbf{y}^\star)\in\mathcal{D}}
\left\|f_{\phi}(\mathbf{x})-\mathbf{y}^\star\right\|_2^2,
\end{equation}
where $\mathcal{D}$ denotes the dataset of load-solution pairs generated by a numerical ACOPF solver.}
The model architecture consists of an input layer, two hidden layers, and an output layer. LG-ND performs densification exclusively on the two hidden layers, while the input and output layer dimensions remain fixed for a given power system topology.

\begin{algorithm}[h]
\caption{LG-ND: Loss Guided Neural Densification}
\label{alg:fg-adp}
\begin{algorithmic}[1]
\Require $\{\Delta,\;\texttt{max\_neurons}\}$
\State Init model with \texttt{minimal} architecture; $\mathcal{L}^\star \gets \infty$ 
\While{True}
    \State Train model until validation loss $\mathcal{L}$ converges 
    \If{$\mathcal{L} \ge \mathcal{L}^\star$} \Return model \Comment{No improvement} 
    \EndIf
    \State $\mathcal{L}^\star \gets \mathcal{L}$ and save current model snapshot 
    \If{Total neurons $\ge \texttt{max\_neurons}$} \Return model 
    \EndIf
    \State Expand each hidden layer by $\Delta$ neurons 
\EndWhile
\end{algorithmic}
\end{algorithm}

\added{The condition $\mathcal{L} \ge \mathcal{L}^\star$ provides a natural stopping rule for densification. It identifies the point at which the network capacity becomes sufficient to represent the ACOPF constraint structure. Expanding model's architecture beyond this threshold leads to an increase the computational overhead for the current optimization problem.
}

\section{Constraint Clipping for Feasibility Analysis}
\rev{In addition to the unconstrained proxy outputs, we also investigate a clipped variant of the model to study the tradeoff between operational feasibility and optimality. While the learned proxy $f_\phi(\mathbf{x})$ approximates the ACOPF solution manifold, unconstrained predictions may violate operational inequality limits.}

\rev{To mitigate this issue, we apply a coordinate-wise clipping operator to constrained output dimensions during inference. For each bounded variable, \(y_i\), the clipped prediction \(\tilde{y}_i\)  
 is computed as:
 \[
\tilde{y}_i = \max\left( \min\left(y_i,\, y_i^{\max}\right),\, y_i^{\min} \right)
\]
where \( y_i^{\min} \) and \( y_i^{\max} \) denote the admissible lower and upper operating limits, respectively.
The clipped and unclipped variants are evaluated separately throughout the experiments to analyze the impact of bounded post-processing on optimality gap and physical feasibility. The resulting tradeoffs are discussed in Section IV.}




\begin{figure*}[t]
    \centering
    \includegraphics[width=\textwidth]{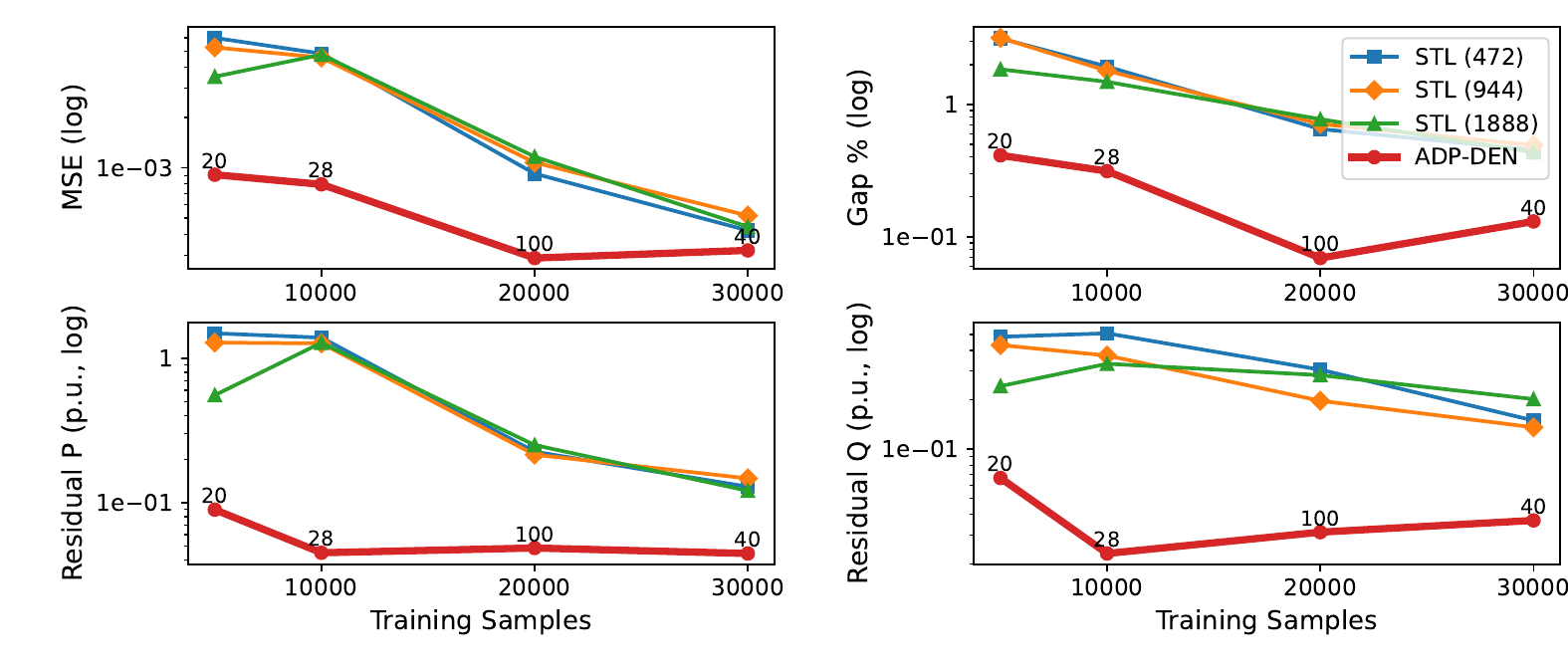}
    \caption{Comparison of STL and ADP-DEN clipped models with their network size on IEEE 118-bus system. Here STL refers to Naïve MSE models with different neural width.}
    \label{fig:118_compare}
\end{figure*}
\begin{table*}[t]
    \centering
    \caption{Comparative performance results using \textbf{Unclipped Models} for the ACOPF Problem for `case118' with 20000 labeled training samples}
    \vspace{1em}
    \small
    \begin{tabular}{lcccccc}
        \hline
        Method & Hidden Layer Size & Gap\% & Max Eq. & Mean Eq. & Max Ineq. & Mean Ineq. \\
        \hline
        MSE + LG-ND         & 50 (2$\times$) &0.0517  & 0.2137   &0.0485 & 0.1799  & 0.0312 \\
        \hline
        Naïve MAE  $\|\cdot\|_1$   & 472 (2$\times$) & 1.1422 & 0.1685 & 0.1292 & 0.0059 & 0.0015  \\
        Naïve MSE $\|\cdot\|_2$    & 472 (2$\times$) & 0.5474 & 1.7957 & 0.2152  & 1.6025 & 0.2024 \\
        MAE + Penalty              & 472 (2$\times$) & 3.8369 & 0.8273 & 0.1513 & 0.8987 & 0.2909 \\
        MSE + Penalty              & 472 (2$\times$) & 1.9063 & 1.1033 & 0.1584 & 0.6904 & 0.1893 \\
        \hline
    \end{tabular}
    \label{tab:case118_unclipped}
    \vspace{-1em}
\end{table*}

\begin{figure*}[!t]
  \centering
\includegraphics[width=\linewidth]{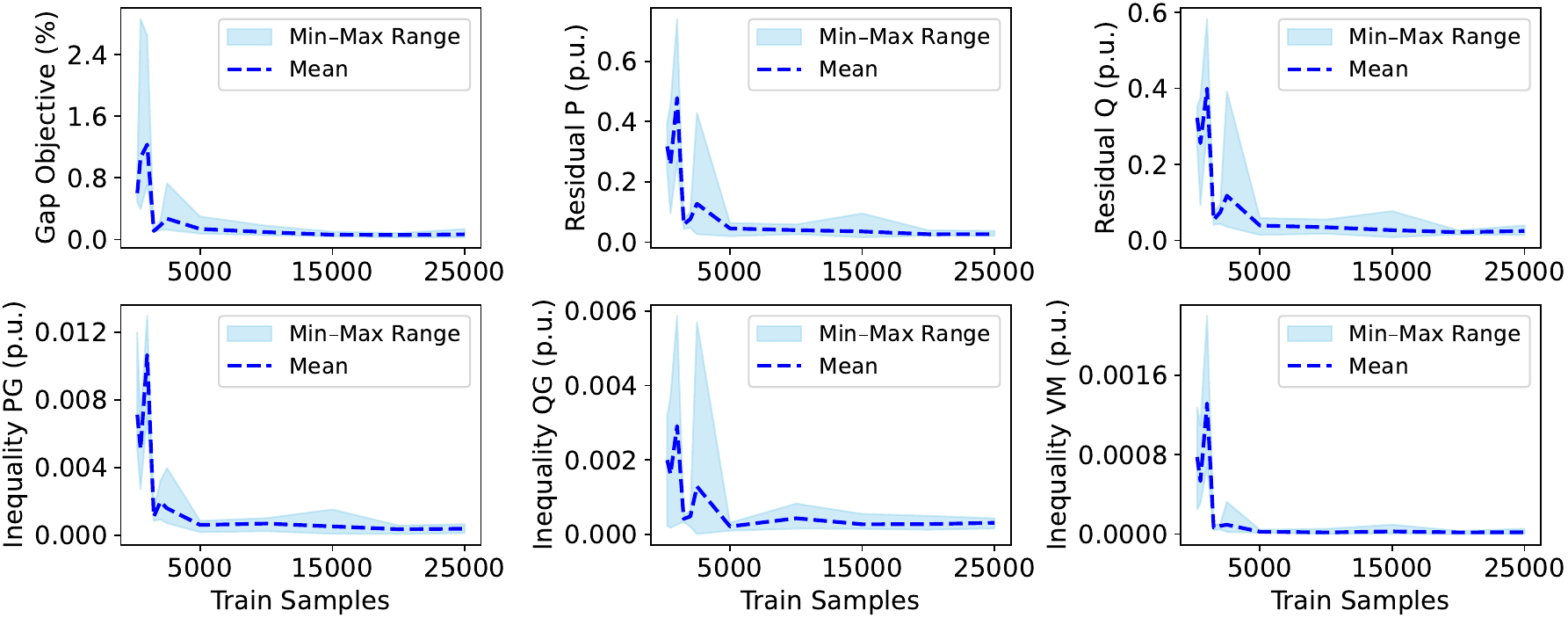}
\vspace{-2em}
  \caption{Robustness and feasibility analysis for the IEEE 57-bus system using the proposed ADP-DEN (LG-ND) framework. The plots illustrate the convergence of the Mean Gap Objective and physical constraint residuals (Real/Reactive Power and Inequality limits) as a function of training sample size. The shaded areas represent the min-max range over multiple training runs, highlighting the stability and high-fidelity approximation of the ACOPF manifold achieved as the network adaptively densifies.}
  \label{fig:57}
\end{figure*}

\vspace*{2cm}
\section{Empirical Results and Discussions}
The proxy performance is evaluated using the standard error matrices: relative optimality gap (Gap\%), equality residuals, and inequality residuals. For both equality and inequality constraints, we report the mean and maximum residuals across the test dataset in per-unit (p.u.). Comparison against various standard ACOPF proxies is carried out.\cite{ML4OPF2023}. 

The proposed LG-ND model demonstrated superior prediction accuracy and optimality compared to standard Naïve MSE (refereed as STL for brevity) and PenaltyMSE approaches on the 57-bus system. The model achieved a Mean MSE of $0.000102$ with only 160 total neurons, representing an 85.7\% reduction in error relative to the STL baseline. Furthermore, the lowest Mean Gap Objective was recorded at 0.0313\% using just 80 neurons. The robustness analysis of the ADP-DEN model for the 57-bus system demonstrates consistent convergence across all metrics as the training volume increases in Figure \ref{fig:57}. As samples scale toward 25000, the Mean Gap Objective stabilizers significantly below 0.5\%. Physical consistency is maintained through the sharp decay in real and reactive power residuals. Notably, the voltage magnitude (VM) residuals reach a precision near $10^{-4}$ p.u., confirming that the adaptive process successfully resolves the nonlinearities of the power flow manifold. 

The LG-ND process demonstrates that massive over-parameterization is unnecessary for accurate ACOPF approximation. As shown in Fig. \ref{fig:118_compare}, the adaptive model converges to a ``lean" architecture of approximately 40--140 total neurons,beyond which additional width yields only marginal gains. 

\begin{itemize}
    \item \textbf{Accuracy and Optimality:} As detailed in Table \ref{tab:case118_unclipped}, the LG-ND model (\texttt{MSE + ADP MSE}) with a hidden layer size of only 50 (2$\times$) achieves a superior Gap\% of 0.0517\%. This significantly outperforms the wider Naïve MSE and Penalty-based baselines\footnote{All baselines models are taken from \cite{ML4OPF2023}, along with the error definitions.} (472 2$\times$), which exhibit gaps ranging from 0.54\% to 1.90\%.
    \item \textbf{Feasibility and Width:} Despite having nearly ten times fewer neurons per layer than standard literature models, the lean proxy maintains highly competitive feasibility. For the 118-bus system, it achieves a Mean Equality residual of 0.0485 p.u., which is nearly five times better than the much larger Naïve MSE baseline (0.2152 p.u.).
    \item \textbf{Verifiability:} Reducing layer width from typical values ($>512$) to strict physical necessity (approx. 50--70 neurons) is a functional requirement for formal safety verification. This reduction makes branch-and-bound verifiers like $\beta$-CROWN \cite{wang2021betacrown} computationally tractable for real-time grid operations.

\end{itemize}
    
\noindent \added{An additional advantage of LG-ND is that the resulting networks are substantially smaller than standard ACOPF proxies. Since formal verifiers such as $\alpha,\beta$-CROWN~\cite{wang2021betacrown} rely on branch-and-bound procedures whose cost grows rapidly with network width, compact architectures are more amenable to certification in safety-critical power system applications}.

\added{
\noindent\textbf{Impact of Constraint Clipping:}
Table~\ref{tab:case118_unclipped} \rev{reports the unconstrained model predictions, while Table~\ref{tab:case118_clipped} shows the corresponding clipped results after enforcing voltage and flow limits. Clipping removes all inequality violations by construction and reduces the mean equality residual from 0.0485 to 0.0399 p.u. This comes with only a small increase in the optimality gap (0.0517\% to 0.0691\%), which is acceptable in safety-critical grid operation where feasibility is prioritized over marginal cost improvements. Among all evaluated approaches, the proposed LG-ND model exhibits the most favorable feasibility–optimality tradeoff under bounded post-processing. Clipping eliminates all inequality violations while improving equality residuals and incurring only marginal degradation in optimality gap. In contrast, several wider baseline models rely more heavily on clipping to correct large infeasibilities, often accompanied by deteriorated equality consistency or substantially larger optimality gaps.}

We further observe that constraint satisfaction improves steadily with more training samples. At low data regimes ($<5000$ samples), residual violations remain noticeable, indicating that sufficient training coverage is necessary to accurately learn the ACOPF operating behavior}.

\rev{The inference complexity of each model is quantified using
the total number of trainable parameters, which serves as a
proxy for floating-point operations (FLOPs) in fully connected
neural networks\cite{goodfellow2016deep}. For an MLP with
input dimension $d_{\mathrm{in}}$, hidden layers $h_1, h_2$,
and output dimension $d_{\mathrm{out}}$, the parameter count is}
\[
P =
(d_{\mathrm{in}}h_1 + h_1)
+
(h_1 h_2 + h_2)
+
(h_2 d_{\mathrm{out}} + d_{\mathrm{out}})
\]

\rev{where each term accounts for the weights and biases between successive layers. Since inference in dense neural architectures is
dominated by matrix multiplication operations, parameter count
is commonly used as a surrogate measure of inference cost.}



\begin{table}[t]
    \centering
    \vspace{-1em}
    \caption{Comparative performance results using \textbf{Clipped Models} for the ACOPF problem on `case118'.}
    \vspace{-0.7em}
    \small
    \begin{tabular}{lccc}
        \hline
        Method & Gap\% & Mean Eq. & Mean. Ineq. \\
        \hline
        MSE + LG-ND ($\mathcal{L}$)  & 0.0691  & 0.03988 & 0.0 \\
        \hline
        Naïve MAE  $\|\cdot\|_1$   & 1.1598 & 0.1253  & 0.0  \\
        Naïve MSE $\|\cdot\|_2$    & 0.7118 & 0.2062 & 0.0  \\
        MAE + Penalty              & 0.8055 & 0.2051 & 0.0  \\
        MSE + Penalty              & 0.9190 & 0.2199 & 0.0 \\
        \hline
        \multicolumn{4}{l}{Bottom four DNN models width: 472 (2x); $N_{train}$ = 20000.}
    \end{tabular}
    \label{tab:case118_clipped}
    \vspace{-1em}
\end{table}

\begin{table}[h]
\centering
\caption{Inference complexity comparison on IEEE-118.}
\label{tab:inference}
\begin{tabular}{lccc}
\hline
Method & Hidden Width & Parameters & Relative Cost \\
\hline
Na\"ive MSE & 472 (2$\times$) & 498K & 15.58$\times$ \\
Na\"ive MAE & 472 (2$\times$) & 498K & 15.58$\times$ \\
Penalty MSE & 472 (2$\times$) & 498K & 15.58$\times$ \\
Penalty MAE & 472 (2$\times$) & 498K & 15.58$\times$ \\
\textbf{LG-ND (Ours)} & \textbf{50 (2$\times$)} & \textbf{31.9K} & \textbf{1.0$\times$} \\
\hline
\end{tabular}
\end{table}

\rev{The proposed LG-ND framework achieves approximately
15$\times$ lower inference complexity compared to conventional
fixed-width ACOPF proxy architectures while maintaining
superior optimality performance.}

\section{Conclusions}
Through the LG-ND process, we have demonstrated that ACOPF proxies can maintain high performance with up to ten times fewer neurons than current literature baselines. This architectural minimalism ensures competitive optimality achieving relative gaps as low as 0.03\% while directly enabling the formal safety verification necessary for real-time grid deployment. Future work will investigate the impact of grid topology perturbations on these lean architectures and integrate them into certifiable neural control loops.

\bibliographystyle{IEEEtran}
\bibliography{main}

\end{document}